\documentclass[journal,twoside,web]{ieeecolor}
\usepackage{tmi}
\usepackage{cite}
\usepackage{amsmath,amssymb,amsfonts}
\usepackage{graphicx}
\usepackage{textcomp}
\usepackage{url}
\usepackage{float}
\usepackage{xcolor}
\usepackage{algpseudocode}
\usepackage{array}
\usepackage[caption=false,font=normalsize,labelfont=sf,textfont=sf]{subfig}
\usepackage{textcomp}
\usepackage{stfloats}
\usepackage{url}
\usepackage{verbatim}
\usepackage{graphicx}
\usepackage{graphbox}
\usepackage{makecell}
\usepackage{tabularx}
\usepackage{bm}
\usepackage{booktabs} 
\usepackage{colortbl} 
\usepackage{xcolor}   
\usepackage[ruled,vlined,linesnumbered]{algorithm2e}
\usepackage{cleveref}
\usepackage{multirow}
\usepackage{booktabs}

\def\BibTeX{{\rm B\kern-.05em{\sc i\kern-.025em b}\kern-.08em
    T\kern-.1667em\lower.7ex\hbox{E}\kern-.125emX}}
\begin{document}
\title{PAINT: Pathology-Aware Integrated Next-Scale Transformation for Virtual Immunohistochemistry}
\author{Rongze Ma, Mengkang Lu, Zhenyu Xiang, Yongsheng Pan, Yicheng Wu, Qingjie Zeng, Yong Xia \IEEEmembership{Member, IEEE}
\thanks{This work was supported in part by the National Natural Science Foundation of China under Grant 92470101 and Grant 62401475.
\textit{(R. Ma and M. Lu contributed equally to this work.) (Corresponding authors: Qingjie Zeng and Yong Xia.)}}
\thanks{R. Ma, M. Lu, Z. Xiang, Y. Pan, Q. Zeng, and Y. Xia are with 
School of Computer Science and Engineering, Northwestern Polytechnical University, Xi’an 710072, China (e-mail: \{rongzema129, lmk, xiangzy, qjzeng\}@mail.nwpu.edu.cn; \{yspan, yxia\}@nwpu.edu.cn). } 
\thanks{Y. Wu is with Monash University, Clayton, VIC 3168, Australia (e-mail: yicheng.wu@monash.edu).}
}

\maketitle

\begin{abstract}

Virtual immunohistochemistry (IHC) aims to computationally synthesize molecular staining patterns from routine Hematoxylin and Eosin (H\&E) images, offering a cost-effective and tissue-efficient alternative to traditional physical staining. However, this task is particularly challenging: H\&E morphology provides ambiguous cues about protein expression, and similar tissue structures may correspond to distinct molecular states.
Most existing methods focus on direct appearance synthesis to implicitly achieve cross-modal generation, often resulting in semantic inconsistencies due to insufficient structural priors.
In this paper, we propose Pathology-Aware Integrated Next-Scale Transformation (PAINT), a visual autoregressive framework that reformulates the synthesis process as a structure-first conditional generation task. Unlike direct image translation, PAINT enforces a causal order by resolving molecular details conditioned on a global structural layout.  Central to this approach is the introduction of a Spatial Structural Start Map (3S-Map), which grounds the autoregressive initialization in observed morphology, ensuring deterministic, spatially aligned synthesis. Experiments on the IHC4BC and MIST datasets demonstrate that PAINT outperforms state-of-the-art methods in structural fidelity and clinical downstream tasks, validating the potential of structure-guided autoregressive modeling.

\end{abstract}


\section{Introduction}
\label{sec:introduction}

Immunohistochemistry (IHC) is a fundamental tool in modern pathological diagnosis, enabling the visualization of protein expression patterns that guide tumor subtyping, prognostic assessment, and decision-making~\cite{gurcan2009histopathological,niazi2019digital}. Despite its clinical importance, IHC remains resource-intensive, requiring additional tissue sections and specialized reagents. These limitations are particularly restrictive when tissue availability is scarce or rapid diagnostic turnaround is required. Consequently, there has been growing interest in virtual IHC, which aims to generate IHC staining patterns directly from routine Hematoxylin and Eosin (H\&E) slides~\cite{rivenson2018deep,ghahremani2022deepliif}.

Virtual IHC poses a fundamentally ill-posed inference problem. While H\&E images encode tissue morphology and cellular organization, IHC reveals molecular expression that is only indirectly reflected in morphology~\cite{lu2023multi,he2020integrating}. The relationship between H\&E morphology and protein expression is inherently ambiguous and underdetermined: morphologically similar regions may correspond to distinct molecular states due to differences in cellular composition, microenvironment, or disease progression~\cite{pspstain}.
This intrinsic uncertainty implies that virtual IHC cannot be treated as a deterministic modality translation task. Instead, it is a conditional generation task that requires hierarchical reasoning, where global structural context must be established before resolving local molecular patterns. This hierarchical reasoning mirrors clinical practice, in which pathologists first interpret coarse architectural cues (e.g., tumor boundaries, glandular organization, and stromal interfaces) to contextualize subsequent assessment of molecular markers at finer scales. Consistently, IHC expression patterns are largely governed by region-level cellular composition and microenvironmental context, which must be inferred at a coarse scale prior to synthesizing fine-grained molecular details under incomplete morphological constraints~\cite{tian2025coarse, srinidhi2021deep}.

Most existing virtual staining methods formulate the task as deterministic image-to-image (I2I) translation, commonly adopting general-purpose frameworks such as Pix2Pix~\cite{pix2pix} and CycleGAN~\cite{cyclegan}.
However, directly transferring these objectives to pathology can be problematic, as pixel-level reconstruction and cycle-consistency losses lack pathology-specific semantic constraints and often prioritize visual realism over biological plausibility. This can lead to \emph{structural hallucination}, where marker positivity or negativity is incorrectly assigned across regions that are visually similar but biologically distinct (e.g., false positivity outside tumor nests), yielding spatially inconsistent yet visually plausible staining patterns ~\cite{cohen2018distribution,bhadra2021hallucinations,tivnan2024hallucination,dayarathna2025mu}.
Recent works have attempted to alleviate this issue by introducing pathology-aware constraints, such as contrastive alignment or semantic preservation losses~\cite{ppt,pspstain}. 
While these strategies improve local correspondence, they remain within a deterministic translation paradigm and primarily regulate appearance or patch-level fidelity, without explicitly enforcing region-level structural consistency between H\&E and IHC. This limitation contrasts with clinical pathology principles, where molecular signals are expected to remain spatially coherent with tissue compartments and architectural organization~\cite{srinidhi2021deep}.

\begin{figure}[t]
  \centering
  \includegraphics[width=\linewidth, page=5]{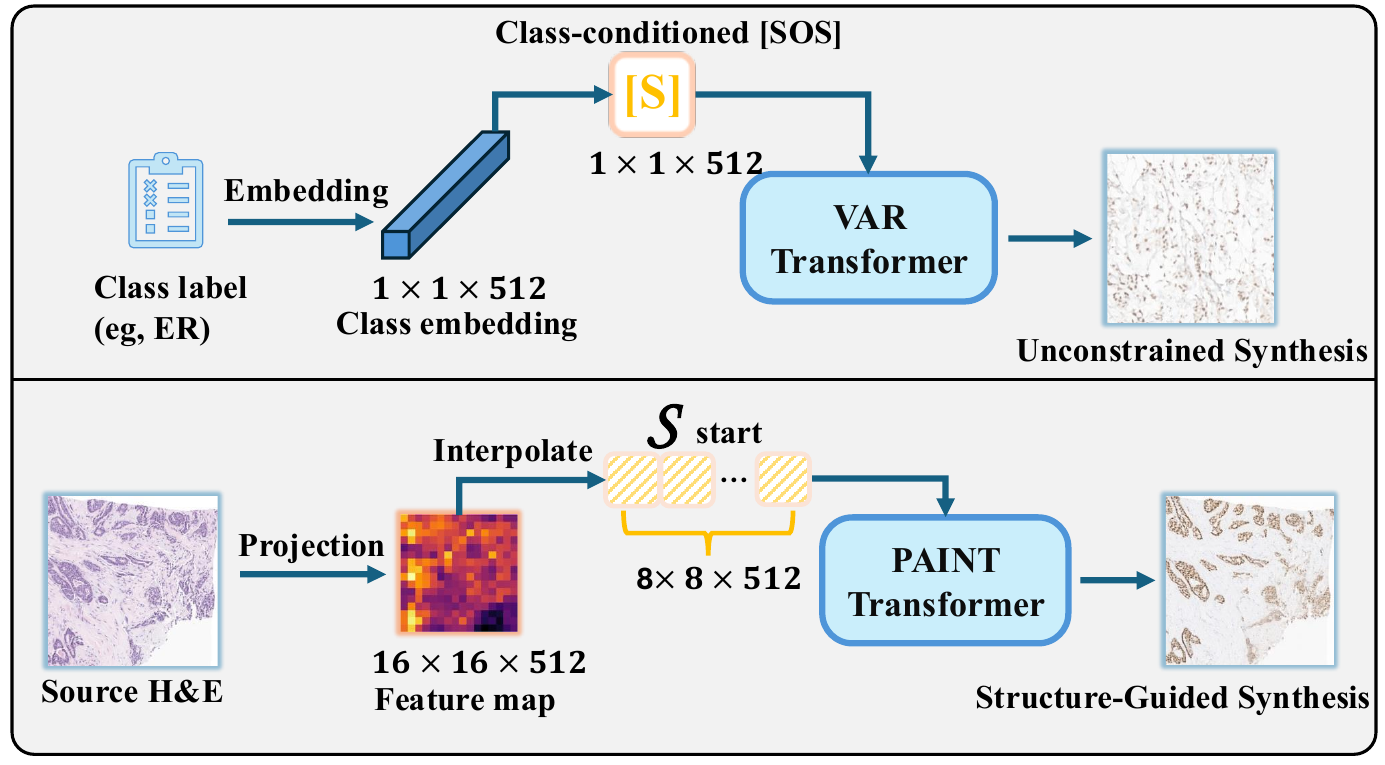} 
  \caption{Standard VAR (top) relies on a generic token (Start of Sequence
[SOS]) lacking morphological grounding. In contrast, PAINT (bottom) introduces a \textbf{Spatial Structural Start Map} ($S_{start}$)—a dense feature prior synthesized from the H\&E image—that strictly aligns the autoregressive synthesis with the source morphology.}
  \label{fig:start_map_comparison}
\end{figure}

These observations motivate a shift from deterministic translation toward hierarchical generative modeling. The recently developed visual autoregressive models (VAR)~\cite{VAR} offer a promising alternative by generating images in a coarse-to-fine manner, first generating low-resolution latent representations and then progressively refining high-resolution details conditioned on this coarse context~\cite{vqvae2}. Such generation order aligns naturally with pathological reasoning, which prioritizes global tissue architecture before focusing on cellular-level interpretation~\cite{srinidhi2021deep}. However, directly applying VAR to pathology images remains challenging. In the absence of explicit spatial anchoring to the source H\&E morphology, early generative decisions lack a morphology-grounded structural hypothesis to condition subsequent refinement, leading to spatial drift and error accumulation during autoregressive synthesis~\cite{bengio2015scheduled} (Fig.~\ref{fig:start_map_comparison}, top). Consequently, ensuring precise spatial alignment and reducing anatomically implausible layouts remains a critical challenge, since the model must infer molecular signals while strictly adhering to the spatial constraints of the source morphology.

To address these challenges, we propose \textbf{P}athology-\textbf{A}ware \textbf{I}ntegrated \textbf{N}ext-Scale \textbf{T}ransformation (PAINT), a VAR framework tailored for virtual IHC. Unlike standard VAR that relies on a generic [SOS] token for class-conditional generation, PAINT introduces a dual-constrained \textbf{Spatial Structural Start Map} (3S-Map) that enforces feature alignment in both the latent space and the image space, providing an IHC-aligned spatial prior for initializing autoregressive synthesis (see Fig.~\ref{fig:start_map_comparison}, bottom). Specifically, PAINT operates in three stages.
First, PAINT separately trains two VQ-VAEs on H\&E and IHC images to learn hierarchical, modality-specific discrete token maps~\cite{wu2025codebrain}, thereby establishing dedicated perceptual vocabularies for tissue morphology and molecular staining patterns. 
Second, PAINT trains a cross-modal Translator that projects frozen H\&E latent representations into the continuous IHC feature space via Latent-Space Alignment (LSA) and Image-Space Alignment (ISA), explicitly enforcing semantic and spatial consistency across modalities. 
Finally, the aligned features are interpolated to construct the \emph{3S-Map}, which initializes the autoregressive generation trajectory and serves as an IHC-aligned spatial prior, guiding subsequent visual autoregressive synthesis toward anatomically coherent layouts while progressively rendering IHC-specific molecular textures.
Through this design, PAINT achieves state-of-the-art (SOTA) performance on the IHC4BC\footnote{\url{https://www.kaggle.com/datasets/akbarnejad1991/ihc4bc-compressed}} and MIST datasets, improving both visual fidelity and clinically relevant prediction accuracy.

Our main contributions are threefold:
\begin{itemize}
  \item We propose \textbf{PAINT}, a pathology-aware VAR framework for virtual IHC that incorporates explicit spatial structural priors into synthesis, enabling pathology-consistent cross-modal generation guided by tissue morphology.
  \item We introduce a \textbf{3S-Map} that is jointly constrained in the latent and image spaces via LSA and ISA, providing an IHC-aligned spatial prior that mitigates region-level structural shift during autoregressive synthesis.
  \item Experiments on the IHC4BC and MIST datasets show that PAINT consistently outperforms existing methods in both image quality and downstream clinical tasks.
\end{itemize}

\section{Related Work} \label{sec:RelatedWork}

\subsection{Generative Models for Image Synthesis and Translation}

I2I translation aims to learn mappings between source and target images while preserving structural content. Pix2Pix~\cite{pix2pix} uses conditional GANs for supervised translation with paired data, but it struggles with medical imaging due to the need for precise pixel-level correspondence. To address this, CycleGAN~\cite{cyclegan} introduces cycle-consistency for unpaired translation. However, CycleGAN’s reliance on strict cycle-consistency can be problematic in medical imaging, where complex or weakly labeled relationships may lead to inaccurate translations and fail to capture nuanced morphological-molecular correlations.
To mitigate these issues, CUT~\cite{cut} utilizes contrastive learning to promote patch-level correspondence, improving weakly-labeled scenarios. Recent diffusion models, including SynDiff~\cite{syndiff} and latent diffusion models (LDMs)~\cite{LDM}, have shown promising results through iterative denoising for improved quality and stability. Despite these advances, these methods continue to treat cross-modal synthesis mainly as style transfer, lacking mechanisms to enforce the global semantic and structural consistency needed for accurate medical interpretations.
Autoregressive generative models, such as VQ-VAEs, model images as sequences of discrete latent tokens, learning compact and semantically meaningful representations that separate high-level structure from low-level appearance~\cite{esser2021taming,yu2021vector,VQI2I}. 
But conventional raster-scan autoregressive generation is computationally inefficient and may disrupt spatial coherence~\cite{parmar2018image,van2016pixel}. 
Recent VAR modeling addresses these limitations through next-scale prediction, generating images progressively from coarse layouts to fine details~\cite{VAR}. By explicitly encoding multi-scale spatial hierarchy, VAR offers a structured and spatially grounded generation strategy.

\subsection{Virtual Staining in Computational Pathology} 

To ensure effective visual staining in pathology, many existing methods introduce pathology-specific constraints.
Specifically, ASP~\cite{ASP-MIST} introduces adaptive PatchNCE loss to reduce sensitivity to alignment issues, while PPT~\cite{ppt} uses bidirectional contrastive learning with patch alignment loss to address misalignment in paired images. However, both focus mainly on patch-level consistency and still struggle with global semantic alignment, especially in regions with severe misalignment. PSPStain~\cite{pspstain} improves molecular specificity by modeling protein expression with optical density estimation and prototype consistency learning. While effective in preserving molecular semantics, its performance depends on the quality of these strategies, particularly when dealing with spatial misalignment. HistDiST~\cite{histdist} applies latent diffusion with dual morphological conditioning to enhance structural and molecular fidelity, but its reliance on base models for inference limits flexibility and fine-tuning for diverse molecular features.
Despite these advances, most virtual IHC staining methods prioritize pixel- or patch-level alignment while neglecting the crucial role of structural priors in capturing spatial relationships among tissue regions. This absence of structure-aware modeling undermines their ability to maintain consistent tissue organization. 

\begin{figure*}[t]
   \centering
   \includegraphics[width=1\textwidth,page=2]{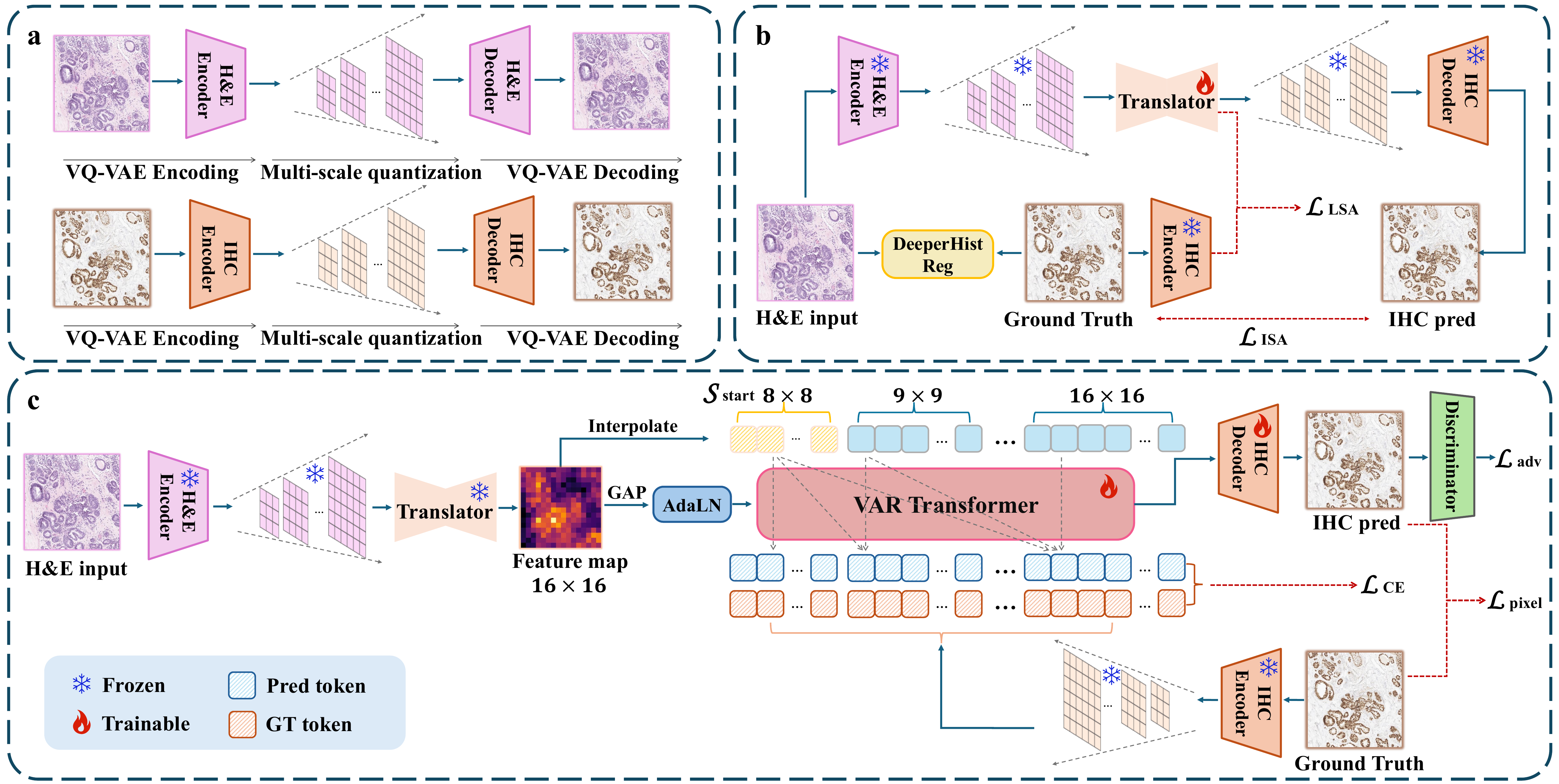}
 \caption{
 Overview of the proposed \textbf{PAINT} framework. (a) PAINT independently trains two domain-specific multi-scale VQ-VAEs for H\&E and IHC images. Each encoder maps the input image to continuous latent features, which are discretized into hierarchical token maps via residual vector quantization. The corresponding decoders reconstruct the images to ensure compact yet high-fidelity representations. After training, all VQ-VAE encoders and decoders are frozen. (b) Using paired H\&E--IHC samples registered by DeeperHistReg, a translation network (U-Net) maps H\&E latent features, extracted by the frozen H\&E encoder, into the continuous IHC feature space. This stage is supervised by LSA for feature consistency and ISA via the frozen IHC decoder for reconstruction fidelity, producing spatially aligned and structurally consistent intermediate representations. (c) A Visual Autoregressive (VAR) Transformer progressively predicts IHC token maps in a coarse-to-fine, next-scale manner. Translated features are injected as structural conditions via Adaptive Layer Normalization (AdaLN), guiding the autoregressive process to preserve tissue morphology while synthesizing realistic molecular textures. The final IHC image is reconstructed by the IHC VQ-VAE decoder.}
   \label{fig:overall}
\end{figure*}

\section{Methodology}
\label{sec:Methodology}

\subsection{Overview}

Fig.~\ref{fig:overall} illustrates the overall PAINT framework, which enables H\&E-to-IHC synthesis by explicitly injecting structural priors into a next-scale transformation architecture. The framework consists of three stages. 
In the first stage, PAINT trains two VQ-VAE models to reconstruct H\&E and IHC images, respectively, aiming to initialize modality-specific representations. 
In the second stage, PAINT learns a feature-level translation from the H\&E latent space to the IHC latent space using paired, spatially registered samples, guided by specially designed ISA and LSA loss constraints to enforce structurally consistent cross-modality correspondences. 
In the final stage, PAINT performs virtual staining through next-scale autoregressive generation, explicitly conditioned on the translated structural prior and a hierarchical generation scheme, thereby anchoring molecular synthesis to patient-specific tissue morphology.


\subsection{Initialization of modality-specific visual perception}

To model the substantial appearance and distributional differences between H\&E and IHC images, PAINT first trains two parallel VQ-VAE models, denoted as $\text{VQ}_{\text{HE}}$ and $\text{VQ}_{\text{IHC}}$. We adopt the multi-scale residual quantization paradigm of VQ-VAE2~\cite{vqvae2}.
Specifically, each image is represented as a pyramid of $K$ discrete latent grids with increasing spatial resolutions, ranging from ($8 \times 8$) to ($16 \times 16$). 
The models $\text{VQ}_{\text{HE}}$ and $\text{VQ}_{\text{IHC}}$ are trained independently with separate codebooks, $\mathcal{Z}_{\text{HE}}$ and $\mathcal{Z}_{\text{IHC}}$, allowing modality-specific appearance statistics to be faithfully preserved. Since the training procedures of $\text{VQ}_{\text{HE}}$ and $\text{VQ}_{\text{IHC}}$ are identical, we describe only the $\text{VQ}_{\text{HE}}$ branch for brevity.
Given an input image $X$, the encoder $\mathcal{E}$ produces a continuous feature map $f = \mathcal{E}(X)$. Multi-scale quantization is concurrently performed in a residual manner. At scale $k$, the residual feature $f_{\mathrm{res}}^{(k)}$ is interpolated to resolution $h_k \times w_k$ and quantized via nearest-neighbor lookup:
\begin{equation}
\begin{aligned}
    r_k &= \mathcal{Q}\big(\text{Interpolate}(f_{\mathrm{res}}^{(k)}, h_k, w_k)\big), \\
    z_k &= \text{lookup}(\mathcal{Z}_{\text{HE}}, r_k),
\end{aligned}
\end{equation}
Here, $\mathcal{Q}(\cdot)$ denotes a nearest-neighbor vector quantizer that, at each spatial location, assigns the interpolated residual feature vector to the index of its closest codeword in the codebook $\mathcal{Z}_{\text{HE}}$ under the Euclidean distance. The residual feature is then updated to preserve information not captured at the current scale:
\begin{equation}
    f_{\mathrm{res}}^{(k+1)} = f_{\mathrm{res}}^{(k)} - \phi_k\big(\text{Interpolate}(z_k, H, W)\big),
\end{equation}
where $f_{\mathrm{res}}^{(1)}$ is initialized as the original encoder feature $f$, $\phi_k$ denotes a scale-specific projection and $(H, W)$ is the resolution of $f$. The aggregated quantized feature map $\hat{f}$ is formed by accumulating the projected features from all scales:
\begin{equation}
    \hat{f} = \sum_{k=1}^{K} \phi_k\big(\text{Interpolate}(z_k, H, W)\big),
    \label{eq:quantized_feature}
\end{equation}


Both $\text{VQ}{_\text{HE}}$ and $\text{VQ}{_\text{IHC}}$ are trained using the bitwise loss~\cite{han2025infinity}, which incorporates reconstruction $\mathcal{L}_{\mathrm{rec}}$, semantic $\mathcal{L}_{\mathrm{feat}}$, perceptual $\mathcal{L}_{\mathrm{P}}$~\cite{lpips}, and realism constraints $\mathcal{L}_{\mathrm{adv}}$~\cite{pix2pix}, formulated as:
\begin{equation}
    \mathcal{L} = 
    \underbrace{\|X - \hat{X}\|_1}_{\mathcal{L}_{\mathrm{rec}}}
    + \underbrace{\|f - \hat{f}\|_1}_{\mathcal{L}_{\mathrm{feat}}}
    + \lambda_p \mathcal{L}_{\mathrm{P}}(X, \hat{X})
    + \lambda_{\mathrm{adv}} \mathcal{L}_{\mathrm{adv}}(X, \hat{X}),
\end{equation}
where $\hat{X} = \mathcal{D}(\hat{f})$ is the reconstructed image produced by the decoder $\mathcal{D}$.


\SetKwInput{KwInput}{Inputs}
\SetKwInput{KwParam}{Hyperparameters}
\SetKwInput{KwReturn}{Return}

\begin{algorithm}[t]
\caption{Multi-scale VQ-VAE Encoding and Reconstruction (H\&E or IHC)}
\label{alg:vqvae_multiscale}

\KwInput{raw image $X$;}
\KwParam{number of scales $K$, resolutions $(h_k, w_k)_{k=1}^K$;}

\BlankLine
\textbf{Encoding: Multi-scale Residual Quantization}

$f = \mathcal{E}(X)$\; $f_{\mathrm{res}}^{(1)} = f$\; $R = [\,]$\;

\For{$k = 1, \cdots, K$}{
    $r_k = \mathcal{Q}\big(\text{Interpolate}(f_{\mathrm{res}}^{(k)}, h_k, w_k)\big)$\;
    $R \leftarrow \text{queue\_push}(R, r_k)$\;
    $z_k = \text{lookup}(\mathcal{Z}, r_k)$\;
    $f_{\mathrm{res}}^{(k+1)} =
    f_{\mathrm{res}}^{(k)} -
    \phi_k\big(\text{Interpolate}(z_k, H, W)\big)$\;
}

\BlankLine
\textbf{Reconstruction: Multi-scale Feature Aggregation}

$\hat{f}^{(1)} = 0$\;

\For{$k = 1, \cdots, K$}{
    $r_k = \text{queue\_pop}(R)$\;
    $z_k = \text{lookup}(\mathcal{Z}, r_k)$\;
    $\hat{f}^{(k+1)} =
    \hat{f}^{(k)} +
    \phi_k\big(\text{Interpolate}(z_k, H, W)\big)$\;
}

\BlankLine
\textbf{Decoding}

$\hat{X} = \mathcal{D}(\hat{f}^{(K+1)})$\;

\KwReturn{reconstructed image $\hat{X}$ ; multi-scale tokens $R$;}
\end{algorithm}

\subsection{Feature Alignment for 3S-Map Initialization}
To prepare the structural prior for H\&E-to-IHC translation, PAINT performs a feature-space mapping from H\&E to IHC using paired images registered with DeeperHistReg~\cite{DeeperHistReg}.
Given an H\&E image $X_{\text{HE}}$, PAINT extracts its aggregated quantized feature map $\hat{f}_{\text{HE}}$ using the pre-trained $\text{VQ}_{\text{HE}}$. 
A translation network $\mathcal{T}$ that utilizes a U-Net backbone~\cite{unet} is then employed to map $\hat{f}_{\text{HE}}$ to a continuous IHC feature prediction:
\begin{equation}
    f_{\text{IHC}}^{\text{pred}} = \mathcal{T}\!\left(\hat{f}_{\text{HE}}\right),
    \label{eq_5}
\end{equation}
where $f_{\text{IHC}}^{\text{pred}}$ is the predicted latent feature from H\&E space.
To achieve the desired structural consistency, two alignment strategies are introduced.

{\noindent\bf{LSA.}} 
To ensure accurate projection from H\&E to IHC in the latent space, we introduce a LSA loss that minimizes the $\ell_1$ distance between $f_{\text{IHC}}^{\text{pred}}$ and $f_{\text{IHC}}^{\text{gt}}$, where $f_{\text{IHC}}^{\text{gt}}$ is obtained by encoding the registered ground-truth IHC image $X_{\text{IHC}}$ using the pre-trained $\text{VQ}_{\text{IHC}}$ encoder. Formally,
\begin{equation}
\label{eq:l_LSA}
    \mathcal{L}_{\text{LSA}} =
    \| f_{\text{IHC}}^{\text{gt}} - f_{\text{IHC}}^{\text{pred}} \|_1,
\end{equation}

{\noindent\bf{ISA.}} 
Beyond feature alignment, we further employ a pixel-level alignment loss, ISA, designed to minimize the $\ell_1$ reconstruction error between the predicted IHC image $\hat{X}_{\text{IHC}}^{\text{pred}}$ and its registered ground-truth counterpart, $X_{\text{IHC}}$. Formally,
\begin{equation}
\label{eq:l_ISA}
    \mathcal{L}_{\text{ISA}} =
    \| X_{\text{IHC}} - \hat{X}_{\text{IHC}}^{\text{pred}} \|_1,
\end{equation}

The Translator $\mathcal{T}$ is jointly trained using both constraints:
\begin{equation}
\label{eq:l_trans}
    \mathcal{L}_{\text{trans}} =
    \mathcal{L}_{\text{LSA}} + \lambda_{\text{trans}} \mathcal{L}_{\text{ISA}},
\end{equation}
where $\lambda_{\text{trans}}$ balances latent-level alignment and image-level spatial fidelity, respectively. 

\subsection{Visual Autoregressive Generation via Structural Conditioning}
Unlike existing autoregressive models that operate unconditionally or rely on weak semantic cues such as class labels~\cite{VAR}, PAINT performs structure-conditioned synthesis to address anatomically implausible layouts in cross-modal generation. Specifically, $f_{\mathrm{IHC}}^{\mathrm{pred}}$, computed according to Eq.~\ref{eq_5}, serves as a spatial anchor, guiding the generation of molecular staining patterns that are strictly aligned with the tissue morphology present in the input H\&E image.
Based on $f_{\mathrm{IHC}}^{\mathrm{pred}}$, two complementary conditioning signals can be derived as expected to guide the generation process: a global context vector for appearance modulation and a spatially grounded start map for geometry-aware autoregressive initialization.

{\noindent\bf{Global Context Formulation.}} 
To capture global representations, global average pooling is applied to $f_{\mathrm{IHC}}^{\mathrm{pred}}$, yielding $f_{\mathrm{global}}$, which is injected into the VAR Transformer via Adaptive Layer Normalization~\cite{huang2017arbitrary,peebles2023scalable}, enabling scale-adaptive modulation of feature statistics during generation.

{\noindent\bf{3S-Map $S_{\mathrm{start}}$.}}
To replace the generic learnable start token (e.g., [SOS]) with a spatially explicit initialization, we construct $S_{\mathrm{start}}$ directly from the Translator output.
Given the aligned feature map $f_{\mathrm{IHC}}^{\mathrm{pred}}$, we downsample it to the resolution of the coarsest scale ($k=1$) to serve as the initial condition for autoregressive modeling. Formally, $S_{\mathrm{start}}$ is computed as:
\begin{equation}
S_{\mathrm{start}}=\operatorname{Interpolate}\!\left(f_{\mathrm{IHC}}^{\mathrm{pred}}, h_1,w_1\right),
\end{equation}
where $(h_1, w_1)$ denotes the spatial dimensions of the coarsest scale within the multi-scale hierarchy depicted in Fig.~\ref{fig:overall}(a). This operation ensures that the generation trajectory is initialized with a deterministic, morphology-grounded structural layout derived from the source tissue architecture.

{\noindent\bf{Next-Scale Autoregressive Modeling.}}
PAINT models multi-scale discrete representations via a hierarchical, structure-conditioned autoregressive process, enabling coarse-to-fine generation under explicit anatomical constraints.
Let $R = (r_1, r_2, \ldots, r_K)$ denote the sequence of discrete token maps across increasing spatial resolutions. The conditional distribution at each scale is defined over all previously generated coarser scales, together with a global structural context extracted from the input:
\begin{equation}
p(R \mid f_{\mathrm{IHC}}^{\mathrm{pred}}) = \prod_{k=1}^{K} p\big(r_k \mid r_{<k}, f_{\mathrm{global}}\big),
\end{equation}
where $r_{<k}$ denotes tokens generated at coarser scales.
To explicitly enforce spatial consistency, the generation is initialized with $r_{<1} = {S_{\mathrm{start}}}$.
This design ensures that each refinement step is jointly conditioned on global semantic context and spatial priors derived from the input H\&E image, yielding anatomically coherent and scale-consistent synthesis.

\subsection{Training Objective and Inference}
{\noindent\bf{Training Objective.}} 
During training, we adopt a teacher-forcing strategy, where ground-truth IHC tokens extracted from the frozen $\text{VQ}_{\mathrm{IHC}}$ supervise the generation via a token-level cross-entropy loss $\mathcal{L}_{\mathrm{CE}}$ applied across all scales.
In parallel, the IHC decoder $\mathcal{D}_{\mathrm{IHC}}^{\mathrm{FT}}$ is jointly fine-tuned to enhance image-level fidelity.
Specifically, the predicted token hierarchy is decoded into an IHC image $\hat{X}_{\mathrm{IHC}}^{\mathrm{pred}}$, which is constrained to match the ground-truth image $X_{\mathrm{IHC}}$ using a combination of pixel-wise reconstruction and adversarial losses.
The complete objective is given by
\begin{equation}
    \mathcal{L}_{\mathrm{AR}} =
    \mathcal{L}_{\mathrm{CE}} +
    \lambda_{\mathrm{1}} \underbrace{\|X_{\mathrm{IHC}} - \hat{X}_{\mathrm{IHC}}^{\mathrm{pred}}\|_1}_{\mathcal{L}_{\mathrm{pixel}}}
    + \lambda_{\mathrm{2}} \mathcal{L}_{\mathrm{adv}}\big(X_{\mathrm{IHC}}, \hat{X}_{\mathrm{IHC}}^{\mathrm{pred}}\big),
\end{equation}
where $\lambda_{1}$ and $\lambda_{2}$ are weighting coefficients.


{\noindent\bf{Inference.}} 
At inference time, the autoregressive process is initialized with the 3S-Map and proceeds to sample discrete tokens scale-by-scale following the learned next-scale distribution. 
The resulting token pyramid $R$ is decoded to produce the synthesized IHC image using the fine-tuned decoder $\mathcal{D}_{\mathrm{IHC}}^{\mathrm{FT}}$.

\section{Experiments}
\label{sec:Experiments}

\subsection{Datasets}
We evaluated our method on two datasets.
\textbf{The IHC4BC dataset}~\cite{IHC4BC} comprises multi-patient breast cancer samples stained for ER, PR, HER2, and Ki67. To prevent data leakage and ensure clinically meaningful evaluation, we adopted a patient-level four-fold stratified cross-validation protocol, with folds balanced according to clinical labels. The dataset includes 59 (ER), 52 (HER2), 60 (PR), and 60 (Ki67) patients, corresponding to 26,134, 20,729, 24,971, and 20,668 patch pairs, respectively.
\textbf{The MIST dataset}~\cite{ASP-MIST} provides aligned H\&E–IHC image pairs for four clinically relevant breast cancer biomarkers, including estrogen receptor (ER), progesterone receptor (PR), HER2, and Ki67, with a fixed official data split. The training sets contain 4,153 (ER), 4,642 (HER2), 4,139 (PR), and 4,361 (Ki67) image pairs, while each test set includes 1,000 paired samples.
 
To ensure reliable pixel-level correspondence between H\&E and IHC modalities, all raw data in both datasets were first registered using DeeperHistReg~\cite{DeeperHistReg}. 


\subsection{Evaluation Metrics}
The quality of generated IHC images was assessed on the test sets using peak signal-to-noise ratio (PSNR)~\cite{jahne2005digital}, structural similarity index measure (SSIM)~\cite{wang2004image}, and learned perceptual image patch similarity (LPIPS)~\cite{lpips}, capturing pixel-level fidelity, structural consistency, and perceptual realism, respectively.


On the IHC4BC dataset, we also evaluated the clinical utility of the generated images on downstream classification and regression tasks. We employed specific evaluation pipelines for each biomarker based on standard pathological scoring guidelines~\cite{wolff2018human,allison2020estrogen}.

For HER2, the evaluation was formulated as a binary classification task, excluding equivocal cases (score 2+) in accordance with clinical guidelines~\cite{wolff2018human}. Samples with scores $\{0,1+\}$ were labeled negative, and those with score $3+$ were labeled positive. Performance was evaluated using Accuracy (ACC) and Area Under the ROC Curve (AUC).


For ER, we evaluated both continuous and categorical clinical scores. Nuclei were first stratified into four intensity levels ($0,1+,2+,3+$) based on diaminobenzidine (DAB) channel intensity thresholds defined by the dataset. Using this stratification, we derived two clinical scores. 
\textit{H-Score (Regression).} The H-Score serves as a weighted measure of staining intensity. To ensure robustness to segmentation errors in the generated IHC domain, the total number of nuclei detected in the original H\&E image ($N_{total}$) was used as the denominator, following the dataset protocol. The H-Score is calculated as:
\begin{equation}
    \text{H-Score} = \sum_{i=1}^{3} \left( i \times \frac{N_{i+}}{N_{total}} \times 100 \right)
\end{equation}
where $N_{i+}$ denotes the count of nuclei with intensity $i+$. The score ranges from 0 to 300. Agreement between predicted and ground-truth H-Scores was quantified using Coefficient of Determination ($R^2$), Spearman's Rank Correlation, Mean Squared Error (MSE), and Pearson Correlation.
\textit{Allred Score (Classification).} The semi-quantitative Allred Score~\cite{ahmad2022allred} is calculated as the sum of the Proportion Score (PS) and the Intensity Score (IS), yielding a total score ranging from 0 to 8. Specifically, the PS discretizes the percentage of positive nuclei ($p_{pos}$) into a six-point scale: 0 ($p_{pos}=0\%$), 1 ($0\% < p_{pos} < 1\%$), 2 ($1\% \le p_{pos} \le 10\%$), 3 ($11\% \le p_{pos} \le 33\%$), 4 ($34\% \le p_{pos} \le 66\%$), and 5 ($p_{pos} \ge 67\%$). The Intensity Score (IS) represents the average staining intensity of the positive cells, categorized as 0 (None), 1 (Weak), 2 (Intermediate), and 3 (Strong). Performance was reported using ACC, AUC, F1-score, and Cohen’s Kappa to reflect both classification accuracy and inter-rater agreement.

For PR and Ki67, we formulated the evaluation of the expression levels as regression tasks. The expression level of PR was evaluated using the H-Score, following the same protocol as ER. Ki67 status was quantified by the percentage of positively stained nuclei relative to the total nuclei count, defined as $\text{Pos}_{\%} = \frac{N_{1+} + N_{2+} + N_{3+}}{N_{total}} $, where all staining intensities ($1+$ to $3+$) are considered positive. For both biomarkers, agreement with ground truth was assessed using $R^2$, Spearman's correlation, MSE, and Pearson correlation.

\subsection{Implementation Details}

\begin{table*}[t]
\centering
\caption{Quantitative comparison on IHC4BC dataset. 
Note: The metric headers (Row 1) apply to the dataset columns grouped below them (Row 2).
\textbf{Bold} indicates best, \underline{underline} indicates second-best.}
\label{tab:ihc4bc_combined_final}

\setlength{\aboverulesep}{0pt}
\setlength{\belowrulesep}{0pt}
\renewcommand{\arraystretch}{1.25} 
\setlength{\tabcolsep}{3.5pt} 

\resizebox{\textwidth}{!}{
    \begin{tabular}{l ccc | ccc}
    
    \toprule
    \multicolumn{1}{l}{\multirow{2}{*}{\textbf{Method}}} 
    & \textbf{SSIM} $\uparrow$ & \textbf{PSNR} $\uparrow$ & \multicolumn{1}{c}{\textbf{LPIPS} $\downarrow$} 
    & \textbf{SSIM} $\uparrow$ & \textbf{PSNR} $\uparrow$ & \textbf{LPIPS} $\downarrow$ \\
    
    \cmidrule(lr){2-4} \cmidrule(lr){5-7} 
    
    \multicolumn{1}{l}{} 
    & \multicolumn{3}{c}{\textbf{IHC4BC-ER}} 
    & \multicolumn{3}{c}{\textbf{IHC4BC-PR}} \\
    
    \midrule
    
    Pix2pix~\cite{pix2pix}   
    & 0.6904 $\pm$ 0.0185 & 27.9732 $\pm$ 1.0074 & 0.3464 $\pm$ 0.0096
    & 0.6907 $\pm$ 0.0215 & 27.5582 $\pm$ 0.8272 & 0.3671 $\pm$ 0.0065 \\
    
    CycleGAN~\cite{cyclegan} 
    & 0.6986 $\pm$ 0.0126 & 26.1295 $\pm$ 1.0702 & 0.3185 $\pm$ 0.0058
    & 0.6972 $\pm$ 0.0211 & 28.1213 $\pm$ 0.9435 & 0.3340 $\pm$ 0.0113 \\
    
    CUT~\cite{cut}         
    & 0.6756 $\pm$ 0.0161 & 25.3983 $\pm$ 0.9927 & 0.3408 $\pm$ 0.0042
    & 0.7067 $\pm$ 0.0182 & 28.8566 $\pm$ 0.6899 & 0.3338 $\pm$ 0.0098 \\
    
    ASP~\cite{ASP-MIST}         
    & 0.7000 $\pm$ 0.0157 & 26.8944 $\pm$ 1.1434 & 0.3234 $\pm$ 0.0041
    & \textbf{0.7342 $\pm$ 0.0175} & 29.8142 $\pm$ 0.6957 & 0.3282 $\pm$ 0.0090 \\
    
    PPT~\cite{ppt}         
    & 0.6699 $\pm$ 0.0194 & 26.5800 $\pm$ 1.2216 & 0.3403 $\pm$ 0.0130
    & \underline{0.7268 $\pm$ 0.0162} & 29.6392 $\pm$ 0.6318 & 0.3398 $\pm$ 0.0099 \\
    
    PSPStain~\cite{pspstain}   
    & 0.6819 $\pm$ 0.0079 & 25.1857 $\pm$ 0.6961 & 0.3292 $\pm$ 0.0047
    & 0.7057 $\pm$ 0.0152 & 28.5074 $\pm$ 0.4618 & 0.3253 $\pm$ 0.0049 \\
    
    SynDiff~\cite{syndiff}     
    & \textbf{0.7284 $\pm$ 0.0134} & 28.5775 $\pm$ 0.7726 & 0.2862 $\pm$ 0.0121
    & 0.7212 $\pm$ 0.0197 & \textbf{31.8711 $\pm$ 0.7466} & 0.3395 $\pm$ 0.0092 \\
    
    PAINT-s 
    & 0.6997 $\pm$ 0.0226 & \textbf{29.2288 $\pm$ 1.4865} & \underline{0.2531 $\pm$ 0.0071}
    & 0.7249 $\pm$ 0.0192 & \underline{31.1782 $\pm$ 0.8128} & \underline{0.2658 $\pm$ 0.0063} \\
    
    PAINT   
    & \underline{0.7015 $\pm$ 0.0207} & \underline{29.1047 $\pm$ 1.4726} & \textbf{0.2362 $\pm$ 0.0049}
    & 0.7233 $\pm$ 0.0211 & 30.8409 $\pm$ 0.8150 & \textbf{0.2608 $\pm$ 0.0087} \\
    
    \midrule[\heavyrulewidth] 
    \addlinespace[3pt]
    
    \multicolumn{1}{l}{} 
    & \multicolumn{3}{c}{\textbf{IHC4BC-HER2}} 
    & \multicolumn{3}{c}{\textbf{IHC4BC-Ki67}} \\
    \cmidrule(lr){2-4} \cmidrule(lr){5-7}
    
    Pix2pix~\cite{pix2pix} 
    & 0.2331 $\pm$ 0.0407 & 14.6460 $\pm$ 0.3559 & 0.5044 $\pm$ 0.0146
    & 0.5468 $\pm$ 0.0946 & 24.9537 $\pm$ 2.2243 & 0.3940 $\pm$ 0.0147 \\
    
    CycleGAN~\cite{cyclegan}   
    & 0.2730 $\pm$ 0.0518 & 14.0350 $\pm$ 0.5465 & 0.4711 $\pm$ 0.0233
    & 0.5560 $\pm$ 0.0897 & 24.6346 $\pm$ 2.4722 & 0.3483 $\pm$ 0.0307 \\
    
    CUT~\cite{cut}         
    & 0.2691 $\pm$ 0.0519 & 14.2630 $\pm$ 0.5410 & 0.4646 $\pm$ 0.0216
    & 0.5211 $\pm$ 0.0830 & 23.5816 $\pm$ 2.0921 & 0.3610 $\pm$ 0.0230 \\
    
    ASP~\cite{ASP-MIST}         
    & 0.2721 $\pm$ 0.0506 & 14.5904 $\pm$ 0.4286 & 0.4619 $\pm$ 0.0200
    & 0.5510 $\pm$ 0.0847 & 24.6968 $\pm$ 2.2736 & 0.3540 $\pm$ 0.0247 \\
    
    PPT~\cite{ppt}         
    & 0.2550 $\pm$ 0.0482 & 14.6189 $\pm$ 0.3877 & 0.4834 $\pm$ 0.0213
    & \underline{0.5589 $\pm$ 0.0791} & 25.1633 $\pm$ 2.0750 & 0.4048 $\pm$ 0.0221 \\
    
    PSPStain~\cite{pspstain}   
    & 0.2702 $\pm$ 0.0493 & 14.4405 $\pm$ 0.4251 & 0.4648 $\pm$ 0.0209
    & 0.5502 $\pm$ 0.0856 & 24.1923 $\pm$ 2.2107 & 0.3542 $\pm$ 0.0259 \\
    
    SynDiff~\cite{syndiff}     
    & \underline{0.2747 $\pm$ 0.0519} & 14.4731 $\pm$ 0.5457 & 0.4688 $\pm$ 0.0228
    & 0.5537 $\pm$ 0.0925 & 24.8316 $\pm$ 2.5527 & 0.3546 $\pm$ 0.0271 \\
    
    PAINT-s 
    & 0.2663 $\pm$ 0.0428 & \textbf{15.4114 $\pm$ 0.2889} & \underline{0.4299 $\pm$ 0.0205}
    & 0.5425 $\pm$ 0.0885 & \textbf{25.7656 $\pm$ 2.3270} & \underline{0.3210 $\pm$ 0.0218} \\
    
    PAINT   
    & \textbf{0.2792 $\pm$ 0.0460} & \underline{15.3933 $\pm$ 0.2714} & \textbf{0.3963 $\pm$ 0.0157}
    & \textbf{0.5592 $\pm$ 0.0771} & \underline{25.3670 $\pm$ 2.1536} & \textbf{0.2944 $\pm$ 0.0239} \\
    
    \bottomrule
    \end{tabular}
}
\end{table*}

All experiments were implemented in PyTorch and conducted on a cluster equipped with four NVIDIA A100 GPUs. High-resolution image patches (e.g., $1024 \times 1024$) were downsampled to $256 \times 256$ for training and evaluation. The VQ-VAE tokenizer employed a codebook of size 4,096 with a patch size of $16 \times 16$. 
The standard PAINT model uses 128 base channels and an embedding dimension of 512, while the lightweight variant PAINT-s adopts 64 base channels and an embedding dimension of 256.
Adversarial training was performed using a modified three-layer PatchGAN discriminator~\cite{pix2pix} that aggregates multi-scale intermediate features to provide richer supervision. Optimization was carried out using Adam~\cite{adam2014method} with an initial learning rate of $1\times10^{-4}$ and a dropout rate of 0.5. Training proceeded in three sequential stages: (a) multi-scale VQ-VAEs were trained for 120 epochs on IHC4BC and 60 epochs on MIST with weights set to $\lambda_{p}=1.0$ and $\lambda_{adv}=0.3$; (b) the cross-modal translation network was trained for 30 epochs on both datasets, utilizing $\lambda_{\text{trans}} = 3.0$; and (c) the VAR Transformer was trained for 200 epochs with $\lambda_{\text{1}}=1.0$ and $\lambda_{\text{2}}=0.3$.


\subsection{Experimental Results}

\subsubsection{Competing Methods}
We compared PAINT with a comprehensive set of baselines, including general-purpose image translation methods (Pix2Pix~\cite{pix2pix}, CycleGAN~\cite{cyclegan}, CUT~\cite{cut}, and SynDiff~\cite{syndiff}) and pathology-specific virtual staining approaches (ASP~\cite{ASP-MIST}, PPT~\cite{ppt}, PSPStain~\cite{pspstain}, and HistDiST~\cite{histdist}). All baselines were evaluated under identical data splits and preprocessing pipelines.




\subsubsection{Results on IHC4BC Dataset}
As presented in Table~\ref{tab:ihc4bc_combined_final}, PAINT demonstrates robust performance and consistently outperforms competing methods across the majority of metrics.
In terms of perceptual quality, PAINT achieves the lowest LPIPS scores across all four biomarkers, recording 0.2362 for ER. This indicates that our autoregressive modeling approach generates images that are perceptually closer to the ground truth, significantly outperforming the advanced diffusion-based SynDiff (0.2862 on ER) and the GAN-based ASP (0.3234 on ER).
Regarding structural and pixel-level fidelity, PAINT consistently ranks within the top two across the majority of SSIM and PSNR evaluations among all competing methods. Even on the IHC4BC-PR subset, PAINT achieves an SSIM score that is highly competitive and comparable to the highest value. These results confirm that PAINT effectively handles the complex staining variations in the IHC4BC dataset, producing high-fidelity images with superior structural details.

\begin{figure*}[t]
    \centering
    
    \newcommand{\sidelabelwidth}{0.03\linewidth}
    
    \newcommand{\imgblockwidth}{0.965\linewidth}
    
    \newcolumntype{Y}{>{\centering\arraybackslash}p{0.0877\linewidth}}

    \setlength{\tabcolsep}{0pt} 
    \offinterlineskip           
    
    \begingroup 
    \normalbaselines 
    \begin{tabular}{p{\sidelabelwidth} YYYYYYYYYYY}
         & 
        \scriptsize \textbf{Input (H\&E)} & 
        \scriptsize \textbf{Pix2Pix} & 
        \scriptsize \textbf{CycleGAN} & 
        \scriptsize \textbf{CUT} & 
        \scriptsize \textbf{ASP} & 
        \scriptsize \textbf{PPT} & 
        \scriptsize \textbf{PSPStain} & 
        \scriptsize \textbf{SynDiff} & 
        \scriptsize \textbf{PAINT-s} & 
        \scriptsize \textbf{PAINT} & 
        \scriptsize \textbf{Real IHC} \\
    \end{tabular}
    \endgroup
    \vspace{2pt} 
    \begin{tabular}{>{\centering\arraybackslash}m{\sidelabelwidth} c}
    
        \rotatebox{90}{\textbf{ER}} & 
        \includegraphics[align=c, width=\imgblockwidth]{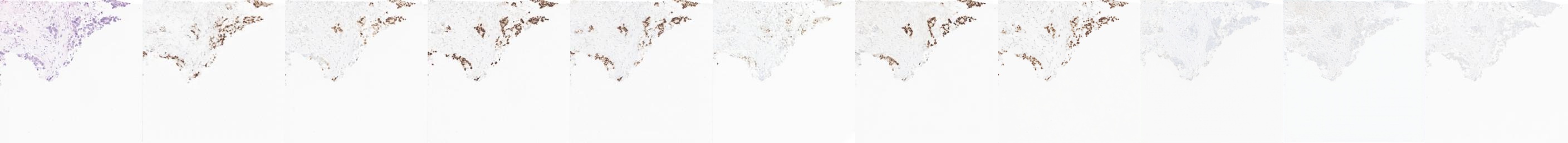} \\
        
        \rotatebox{90}{\textbf{HER2}} & 
        \includegraphics[align=c, width=\imgblockwidth]{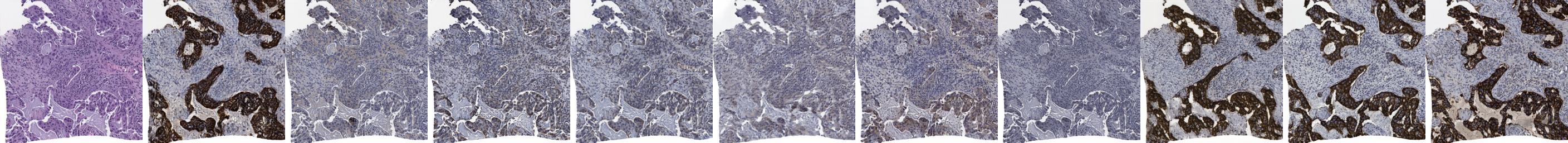} \\
        
        \rotatebox{90}{\textbf{Ki67}} & 
        \includegraphics[align=c, width=\imgblockwidth]{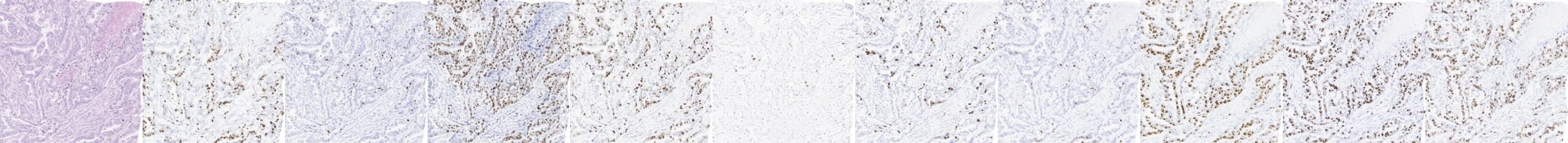} \\
        
        \rotatebox{90}{\textbf{PR}} & 
        \includegraphics[align=c, width=\imgblockwidth]{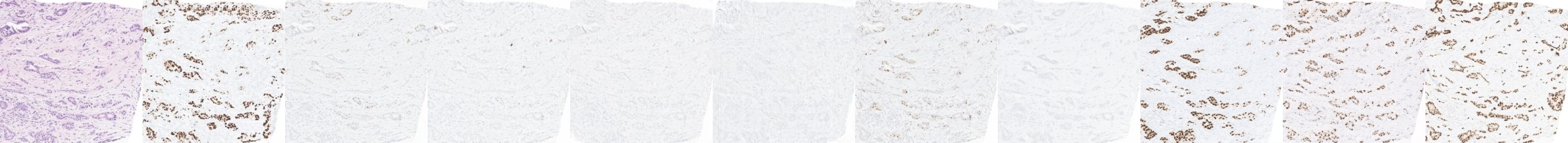} \\
        
    \end{tabular}
    
    \caption{Visual comparison of ER, HER2, Ki67 and PR generation. From top to bottom, the rows represent different IHC markers. The leftmost column shows the input H\&E, and the rightmost column shows the Real IHC.}
    \label{fig:multi_marker_comparison}
\end{figure*}

\subsubsection{Results on Downstream Clinical Tasks.}
We further evaluated the clinical utility of the generated IHC images on downstream diagnostic tasks using the IHC4BC dataset~\cite{IHC4BC}. Specifically, we utilized the UNI-2~\cite{UNI} to extract representations from both the predicted IHC images and ground truth H\&E slides for all downstream tasks. 

{\noindent\bf{HER2 Classification.}} 
As shown in Table~\ref{tab:HER2}, PAINT achieves SOTA performance with an ACC of 0.9753 and an AUC of 0.9902, substantially outperforming all competing methods, such as Pix2Pix (ACC: 0.8948) and CycleGAN (ACC: 0.8093). This demonstrates that PAINT reliably captures the membrane staining patterns necessary for accurate HER2 status determination.

\begin{table}[htbp]
\centering
\caption{Quantitative classification performance on the IHC4BC-HER2+HE dataset. 
\textbf{Bold} indicates the best results, and \underline{underline} indicates the second-best.}

\renewcommand{\arraystretch}{1.1}
\setlength{\tabcolsep}{6pt}

\begin{tabular}{l cc}
\toprule
\textbf{Method} & \textbf{ACC} $\uparrow$ & \textbf{AUC} $\uparrow$ \\
\midrule
H\&E          
& 0.8025 $\pm$ 0.0679 
& 0.8436 $\pm$ 0.0987 \\
Pix2pix~\cite{pix2pix}      
& \underline{0.8948 $\pm$ 0.0386} 
& \underline{0.9370 $\pm$ 0.0203} \\
CycleGAN~\cite{cyclegan}       
& 0.8093 $\pm$ 0.0777 
& 0.8414 $\pm$ 0.1076 \\
CUT~\cite{cut}            
& 0.8118 $\pm$ 0.0693 
& 0.8592 $\pm$ 0.0786 \\
ASP~\cite{ASP-MIST}             
& 0.8206 $\pm$ 0.0569 
& 0.8710 $\pm$ 0.0542 \\
PPT~\cite{ppt}            
& 0.8103 $\pm$ 0.0736 
& 0.8541 $\pm$ 0.0913 \\
PSPStain~\cite{pspstain}       
& 0.8350 $\pm$ 0.0633 
& 0.8814 $\pm$ 0.0555 \\
SynDiff~\cite{syndiff}        
& 0.8033 $\pm$ 0.0746 
& 0.8471 $\pm$ 0.0968 \\
PAINT-s     
& 0.8709 $\pm$ 0.0420 
& 0.9212 $\pm$ 0.0301 \\
PAINT       
& \textbf{0.9753 $\pm$ 0.0130} 
& \textbf{0.9902 $\pm$ 0.0072} \\
\bottomrule
\label{tab:HER2}
\end{tabular}
\end{table}

{\noindent\bf{ER Quantification.}}
For ER assessment, PAINT shows the strongest agreement with ground-truth H-Scores, achieving a Pearson correlation of 0.8664 and an $R^2$ of 0.7354 (Table~\ref{tab:merged_ER_metrics}). In contrast, conventional GANs like CycleGAN and CUT struggle to capture the correct staining intensity, yielding significantly lower $R^2$ values. In Allred Score evaluation, PAINT also attains the highest accuracy for both proportion (0.6313) and intensity (0.7635) components, surpassing the second-best method Pix2Pix by significant margins. This confirms that our coarse-to-fine generation accurately captures both the spatial distribution (proportion) and the pixel-level depth (intensity) of nuclear stains.

\begin{table*}[htbp]
\centering
\caption{Quantitative comparison on IHC4BC-ER. Top: Metric categories. Bottom: Specific metrics. \textbf{Bold} indicates best, \underline{underline} indicates second-best.}
\label{tab:merged_ER_metrics}

\setlength{\aboverulesep}{0pt}
\setlength{\belowrulesep}{0pt}
\renewcommand{\arraystretch}{1.25} 
\setlength{\tabcolsep}{2.5pt}       

\resizebox{\textwidth}{!}{
    \begin{tabular}{l | cccc | cc | cc | cc}
    \toprule
    
    \multicolumn{1}{l}{\multirow{2}{*}{\textbf{Method}}} 
    & \multicolumn{4}{c}{\textbf{H\_Score}} 
    & \multicolumn{2}{c}{\textbf{Allred Score}} 
    & \multicolumn{2}{c}{\textbf{PS}} 
    & \multicolumn{2}{c}{\textbf{IS}} \\
    
    \cmidrule(lr){2-5} \cmidrule(lr){6-7} \cmidrule(lr){8-9} \cmidrule(lr){10-11}
    
    \multicolumn{1}{l}{} 
    & \textbf{R$^2$} $\uparrow$ & \textbf{Spearman} $\uparrow$ & \textbf{MSE} $\downarrow$ & \multicolumn{1}{c}{\textbf{Pearson} $\uparrow$} 
    & \textbf{ACC} $\uparrow$ & \multicolumn{1}{c}{\textbf{AUC} $\uparrow$} 
    & \textbf{ACC} $\uparrow$ & \multicolumn{1}{c}{\textbf{AUC} $\uparrow$} 
    & \textbf{ACC} $\uparrow$ & \textbf{AUC} $\uparrow$ \\
    
    \midrule
    
    H\&E 
    & 0.5292 $\pm$ 0.1895 & 0.7455 $\pm$ 0.1415 & 919.5912 $\pm$ 160.8452 & 0.7651 $\pm$ 0.0830 
    & 0.4317 $\pm$ 0.0153 & 0.8297 $\pm$ 0.0203 
    & 0.5979 $\pm$ 0.0370 & 0.8374 $\pm$ 0.0396 
    & 0.6582 $\pm$ 0.0612 & 0.8191 $\pm$ 0.0280 \\
    
    Pix2Pix~\cite{pix2pix} 
    & 0.6572 $\pm$ 0.1584 & 0.7963 $\pm$ 0.0914 & 654.4841 $\pm$ 86.5599 & 0.8186 $\pm$ 0.0850 
    & 0.4430 $\pm$ 0.0238 & 0.8492 $\pm$ 0.0209 
    & 0.5768 $\pm$ 0.0397 & 0.8335 $\pm$ 0.0424 
    & 0.7088 $\pm$ 0.0371 & \underline{0.8583 $\pm$ 0.0277} \\
    
    CycleGAN~\cite{cyclegan}  
    & 0.3772 $\pm$ 0.2277 & 0.6875 $\pm$ 0.1249 & 1238.1280 $\pm$ 207.0600 & 0.6750 $\pm$ 0.0914 
    & 0.3956 $\pm$ 0.0264 & 0.7998 $\pm$ 0.0067 
    & 0.5489 $\pm$ 0.0403 & 0.8081 $\pm$ 0.0246 
    & 0.6344 $\pm$ 0.0548 & 0.8020 $\pm$ 0.0197 \\
    
    CUT~\cite{cut}  
    & 0.2325 $\pm$ 0.2871 & 0.6271 $\pm$ 0.1255 & 1527.9083 $\pm$ 299.2731 & 0.5956 $\pm$ 0.1043 
    & 0.3791 $\pm$ 0.0288 & 0.7851 $\pm$ 0.0130 
    & 0.5205 $\pm$ 0.0393 & 0.7864 $\pm$ 0.0310 
    & 0.6238 $\pm$ 0.0375 & 0.7964 $\pm$ 0.0264 \\
    
    ASP~\cite{ASP-MIST}  
    & 0.2917 $\pm$ 0.2723 & 0.6566 $\pm$ 0.1300 & 1404.0151 $\pm$ 248.8037 & 0.6340 $\pm$ 0.1021 
    & 0.3818 $\pm$ 0.0162 & 0.7880 $\pm$ 0.0052 
    & 0.5329 $\pm$ 0.0394 & 0.7943 $\pm$ 0.0308 
    & 0.6166 $\pm$ 0.0401 & 0.7918 $\pm$ 0.0187 \\
    
    PPT~\cite{ppt}  
    & 0.1066 $\pm$ 0.2895 & 0.5723 $\pm$ 0.0983 & 1813.8149 $\pm$ 450.2535 & 0.5103 $\pm$ 0.0889 
    & 0.3437 $\pm$ 0.0275 & 0.7349 $\pm$ 0.0115 
    & 0.4879 $\pm$ 0.0170 & 0.7486 $\pm$ 0.0266 
    & 0.5642 $\pm$ 0.0531 & 0.7406 $\pm$ 0.0239 \\
    
    PSPStain~\cite{pspstain}  
    & 0.3860 $\pm$ 0.2184 & 0.6981 $\pm$ 0.1046 & 1243.4932 $\pm$ 346.5706 & 0.6751 $\pm$ 0.0923 
    & 0.4000 $\pm$ 0.0258 & 0.7918 $\pm$ 0.0138 
    & 0.5503 $\pm$ 0.0344 & 0.7985 $\pm$ 0.0226 
    & 0.6536 $\pm$ 0.0524 & 0.8175 $\pm$ 0.0241 \\
    
    SynDiff~\cite{syndiff}  
    & 0.4914 $\pm$ 0.2109 & 0.7356 $\pm$ 0.1120 & 988.7520 $\pm$ 123.9860 & 0.7429 $\pm$ 0.0921 
    & 0.4142 $\pm$ 0.0136 & 0.8192 $\pm$ 0.0127 
    & 0.5776 $\pm$ 0.0304 & 0.8260 $\pm$ 0.0281 
    & 0.6510 $\pm$ 0.0474 & 0.8132 $\pm$ 0.0140 \\
    
    PAINT-s 
    & \underline{0.6991 $\pm$ 0.0983} & \underline{0.8177 $\pm$ 0.0808} & \underline{617.4735 $\pm$ 199.8048} & \underline{0.8427 $\pm$ 0.0557} 
    & \underline{0.4551 $\pm$ 0.0207} & \underline{0.8459 $\pm$ 0.0203} 
    & \underline{0.6103 $\pm$ 0.0460} & \underline{0.8523 $\pm$ 0.0336} 
    & \underline{0.7159 $\pm$ 0.0617} & 0.8510 $\pm$ 0.0083 \\
    
    PAINT 
    & \textbf{0.7354 $\pm$ 0.0496} & \textbf{0.8321 $\pm$ 0.0782} & \textbf{572.8870 $\pm$ 242.1100} & \textbf{0.8664 $\pm$ 0.0261} 
    & \textbf{0.4920 $\pm$ 0.0358} & \textbf{0.8699 $\pm$ 0.0065} 
    & \textbf{0.6313 $\pm$ 0.0286} & \textbf{0.8624 $\pm$ 0.0342} 
    & \textbf{0.7635 $\pm$ 0.0677} & \textbf{0.8979 $\pm$ 0.0250} \\
    
    real IHC 
    & 0.8952 $\pm$ 0.0404 & 0.8870 $\pm$ 0.0715 & 214.3521 $\pm$ 80.0639 & 0.9470 $\pm$ 0.0211 
    & 0.6445 $\pm$ 0.0231 & 0.9395 $\pm$ 0.0094 
    & 0.7391 $\pm$ 0.0301 & 0.9262 $\pm$ 0.0165 
    & 0.9062 $\pm$ 0.0343 & 0.9766 $\pm$ 0.0090 \\
    
    \bottomrule
    \end{tabular}
}
\end{table*}

{\noindent\bf{PR and Ki67 Quantification.}}
The capability of PAINT is further demonstrated on the PR and Ki67. In the challenging PR dataset (see Table~\ref{tab:merged_pr_ki67}), where most competitive methods (including CycleGAN, CUT, and SynDiff) failed to establish a predictive correlation (yielding negative $R^2$ values), PAINT maintained a strong linear relationship with a Pearson correlation of 0.7739 and a positive $R^2$ of 0.5232. This demonstrates that our model correctly infers nuclear expression levels even in cases with subtle staining variances where other generative models collapse. Finally, for the Ki67 proliferation index, which requires precise counting of positive nuclei, PAINT achieves the highest Pearson correlation of 0.8414 and the lowest MSE (0.0017), highlighting its robustness in tasks requiring accurate counting of positive nuclei.

\begin{table*}[t]
\centering
\caption{Quantitative comparison on IHC4BC-PR H\_Score and IHC4BC-Ki67 Positive Percentage prediction. \textbf{Bold} indicates best, \underline{underline} indicates second-best.}
\label{tab:merged_pr_ki67}

\setlength{\aboverulesep}{0pt}
\setlength{\belowrulesep}{0pt}
\renewcommand{\arraystretch}{1.25}
\setlength{\tabcolsep}{2pt}

\resizebox{\textwidth}{!}{
    \begin{tabular}{l | cccc | cccc}
    \toprule
    \multicolumn{1}{l}{\multirow{2}{*}{\textbf{Method}}} & \multicolumn{4}{c}{\textbf{IHC4BC-PR H\_Score}} & \multicolumn{4}{c}{\textbf{IHC4BC-Ki67 Positive Percentage}} \\
    
    \cmidrule(lr){2-5} \cmidrule(lr){6-9}
    
     \multicolumn{1}{l}{} & \textbf{R$^2$} $\uparrow$ & \textbf{Spearman} $\uparrow$ & \textbf{MSE} $\downarrow$ & \multicolumn{1}{c}{\textbf{Pearson} $\uparrow$} & \textbf{R$^2$} $\uparrow$ & \textbf{Spearman} $\uparrow$ & \textbf{MSE} $\downarrow$ & \textbf{Pearson} $\uparrow$ \\
    \midrule
    
    H\&E
    & -0.0994 $\pm$ 0.2295 & 0.4057 $\pm$ 0.1223 & 875.9404 $\pm$ 440.1256 & 0.3830 $\pm$ 0.0725
    & 0.5970 $\pm$ 0.2293 & 0.7903 $\pm$ 0.1335 & 0.0020 $\pm$ 0.0010 & 0.8163 $\pm$ 0.0759 \\
    
    Pix2Pix~\cite{pix2pix}
    & 0.3807 $\pm$ 0.2688 & 0.4905 $\pm$ 0.1083 & \underline{444.4576 $\pm$ 136.0118} & 0.6765 $\pm$ 0.1008
    & 0.4587 $\pm$ 0.3458 & 0.7565 $\pm$ 0.1249 & 0.0026 $\pm$ 0.0011 & 0.7488 $\pm$ 0.1044 \\
    
    CycleGAN~\cite{cyclegan}
    & -0.0299 $\pm$ 0.1911 & 0.4115 $\pm$ 0.0982 & 845.8080 $\pm$ 497.0253 & 0.3723 $\pm$ 0.0728
    & 0.5165 $\pm$ 0.3333 & 0.7769 $\pm$ 0.1317 & 0.0023 $\pm$ 0.0010 & 0.7719 $\pm$ 0.1187 \\
    
    CUT~\cite{cut}
    & -0.0420 $\pm$ 0.2366 & 0.3909 $\pm$ 0.0935 & 842.6920 $\pm$ 480.2599 & 0.3571 $\pm$ 0.0651
    & 0.4663 $\pm$ 0.3644 & 0.7655 $\pm$ 0.1328 & 0.0025 $\pm$ 0.0011 & 0.7592 $\pm$ 0.1009 \\
    
    ASP~\cite{ASP-MIST}
    & -0.0981 $\pm$ 0.2605 & 0.3727 $\pm$ 0.0916 & 886.2112 $\pm$ 505.5282 & 0.3256 $\pm$ 0.0721
    & 0.4979 $\pm$ 0.3447 & 0.7692 $\pm$ 0.1319 & 0.0023 $\pm$ 0.0010 & 0.7750 $\pm$ 0.0976 \\
    
    PPT~\cite{ppt}
    & -0.0405 $\pm$ 0.1780 & 0.3455 $\pm$ 0.0961 & 846.6899 $\pm$ 470.3499 & 0.2923 $\pm$ 0.0536
    & 0.1283 $\pm$ 0.7869 & 0.7124 $\pm$ 0.1198 & 0.0035 $\pm$ 0.0010 & 0.6841 $\pm$ 0.0950 \\
    
    PSPStain~\cite{pspstain}
    & 0.1501 $\pm$ 0.1844 & 0.4771 $\pm$ 0.0856 & 692.5934 $\pm$ 410.0836 & 0.5004 $\pm$ 0.0392
    & 0.4806 $\pm$ 0.3401 & 0.7576 $\pm$ 0.1475 & 0.0025 $\pm$ 0.0011 & 0.7668 $\pm$ 0.0873 \\
    
    SynDiff~\cite{syndiff}
    & -0.0925 $\pm$ 0.2272 & 0.4054 $\pm$ 0.0938 & 881.4252 $\pm$ 475.0513 & 0.3545 $\pm$ 0.0731
    & 0.5416 $\pm$ 0.2973 & \textbf{0.7890 $\pm$ 0.1271} & \underline{0.0022 $\pm$ 0.0010} & 0.7854 $\pm$ 0.1031 \\
    
    PAINT-s
    & \underline{0.3872 $\pm$ 0.2341} & \underline{0.5593 $\pm$ 0.0915} & 453.2665 $\pm$ 170.3116 & \underline{0.6936 $\pm$ 0.0627}
    & \underline{0.5612 $\pm$ 0.2228} & 0.7796 $\pm$ 0.1170 & 0.0023 $\pm$ 0.0011 & \underline{0.7906 $\pm$ 0.0741} \\
    
    PAINT
    & \textbf{0.5232 $\pm$ 0.1965} & \textbf{0.6006 $\pm$ 0.0802} & \textbf{379.2940 $\pm$ 239.0373} & \textbf{0.7739 $\pm$ 0.0526}
    & \textbf{0.6819 $\pm$ 0.1463} & \underline{0.7833 $\pm$ 0.1593} & \textbf{0.0017 $\pm$ 0.0008} & \textbf{0.8414 $\pm$ 0.0626} \\
    
    real IHC
    & 0.8732 $\pm$ 0.0372 & 0.7152 $\pm$ 0.0699 & 107.1590 $\pm$ 79.9160 & 0.9428 $\pm$ 0.0155
    & 0.8819 $\pm$ 0.0697 & 0.8338 $\pm$ 0.1825 & 0.0006 $\pm$ 0.0002 & 0.9452 $\pm$ 0.0325 \\
    
    \bottomrule
    \end{tabular}
}
\end{table*}

\subsubsection{Results on MIST Dataset}
As shown in Table~\ref{tab:mist_results}, PAINT consistently outperforms all competing methods across these biomarkers. Notably, PAINT achieves the highest PSNR for all markers, with significant improvements on PR (16.3104) and Ki67 (16.0698), alongside the best or second-best SSIM, including top performance on the PR (0.2326) and HER2 (0.1983) subsets.
In terms of perceptual quality, PAINT also demonstrates superior performance with the lowest LPIPS scores across all biomarkers, achieving 0.4865 for HER2, surpassing the competitive baseline ASP (0.5021) by a clear margin. These findings validate the effectiveness of PAINT's autoregressive strategy, which significantly reduces blurring and structural hallucinations, ensuring high-quality generation for both nuclear and membrane staining patterns.

\subsubsection{Ablation Study}
Table~\ref{tab:ablation_PAINT} summarizes the impact of key designs in PAINT on the ER-IHC4BC dataset (fold 4), including VAR refinement, alignment losses, decoder adaptation, registration, and the multi-scale autoregressive schedule. The following analysis provides detailed insights into these effects.

{\noindent\bf{Effect of VAR Refinement.}}
Removing the VAR-based refinement stage (\emph{w/o VAR}) reduces PAINT to directly decoding the Translator output with decoder fine-tuning, but without the Visual autoregressive refinement step. Compared to the full PAINT model with VAR, performance decreases, with SSIM dropping from 0.7321 to 0.6668, and PSNR decreasing from 31.2594 to 13.9647. These results underscore the importance of the refinement stage in improving reconstruction fidelity and perceptual quality.

{\noindent\bf{Effect of $\mathcal{L}_{\text{LSA}}$ and $\mathcal{L}_{\text{ISA}}$.}}
Both objectives are involved in shaping the 3S-Map $S_{\text{start}}$.
Specifically, enabling $\mathcal{L}_{\text{LSA}}$ dramatically improves SSIM from 0.6109 to 0.7321 and PSNR from 19.8506 to 31.2594, while reducing LPIPS from 0.3171 to 0.2402.
Similarly, introducing $\mathcal{L}_{\text{ISA}}$ increases SSIM from 0.6046 to 0.7321 and PSNR from 17.9799 to 31.2594, with a corresponding drop in LPIPS from 0.3293 to 0.2402.
These results indicate that incorporating spatial alignment cues into $S_{\text{start}}$ substantially enhances generation quality by strengthening spatial correspondence between H\&E and IHC.

\definecolor{graybg}{gray}{0.95}
\begin{table}[H]
\centering
\caption{Ablation study of PAINT on the ER-IHC4BC dataset (fold4). We investigate the impact of different loss components and scale configurations. The best results are highlighted in \textbf{bold}.}
\label{tab:ablation_PAINT}
\resizebox{\linewidth}{!}{
    \begin{tabular}{l ccc}
    \toprule
    \textbf{Method / Configuration} & \textbf{SSIM} $\uparrow$ & \textbf{PSNR} $\uparrow$ & \textbf{LPIPS} $\downarrow$ \\
    \midrule
    \multicolumn{4}{l}{\textit{\textbf{Component Effectiveness}}} \\
    PAINT w/o VAR & 0.6668 & 13.9647 & 0.4674 \\
    PAINT w/o $\mathcal{L}_{LSA}$ & 0.6109 & 19.8506 & 0.3171 \\
    PAINT w/o $\mathcal{L}_{ISA}$ & 0.6046 & 17.9799 & 0.3293 \\
    PAINT w/o $\mathcal{D}_{\text{IHC}}^{FT}$ & 0.6571 & 18.5512 & 0.4507 \\
    PAINT w/o $\mathcal{L}_{adv}$ & 0.6829 & 22.8388 & 0.3546 \\
    PAINT w/o $\mathcal{L}_{pixel}$ & 0.6765 & 27.0371 & 0.2801 \\
    PAINT w/o $f_{global}$ & 0.7128 & 30.7752 & 0.2493 \\
    PAINT w/o Multi-scale & 0.7078 & 26.7970 & 0.3718 \\
    PAINT w/o Registration & 0.6726 & 29.6429 &  0.3104 \\
    \midrule
    \multicolumn{4}{l}{\textit{\textbf{Scale Configurations}} (with all components enabled)} \\
    Scales: (1-16 dense) & 0.7236 & 31.0121 & 0.2466 \\
    Scales: (1, 2, 4, 8, 16) & 0.6832 & 28.2692 & 0.3322 \\
    \rowcolor{graybg}
    \textbf{PAINT}: (8-16) & \textbf{0.7321} & \textbf{31.2594} & \textbf{0.2402} \\
    
    \bottomrule
    \addlinespace[1pt]
    \multicolumn{4}{l}{\footnotesize \textit{Note: In the ``w/o $\mathcal{D}_{\text{IHC}}^{FT}$'' variant, the decoder is frozen.}} \\
    \end{tabular}
}
\end{table}

{\noindent\bf{Effect of Decoder Fine-tuning and Loss Design.}}
The contribution of task-specific decoder adaptation and loss objectives is validated by the ablations on $\mathcal{D}_{\text{IHC}}^{FT}$, $\mathcal{L}_{adv}$, and $\mathcal{L}_{pixel}$. Freezing the IHC decoder (\emph{w/o $\mathcal{D}_{\text{IHC}}^{FT}$}) in Fig.~\ref{fig:overall}(c) leads to inferior performance (SSIM: 0.6571; PSNR: 18.5512; LPIPS: 0.4507), demonstrating that decoder adaptation is critical for faithfully rendering the learned latent representation into realistic IHC appearance. In addition, removing $\mathcal{L}_{adv}$ noticeably degrades synthesis quality (SSIM: 0.6829; PSNR: 22.8388), indicating that adversarial supervision is essential for enhancing realism and reducing perceptual discrepancy. Similarly, removing $\mathcal{L}_{pixel}$ causes a clear drop in reconstruction fidelity (SSIM: 0.6765; PSNR: 27.0371), suggesting that pixel-wise regression provides a strong low-level anchor for stable synthesis.

{\noindent\bf{Effect of Global Conditioning.}}
The global conditioning signal $f_{global}$ provides patch-level contextual guidance and is injected into the generator via AdaLN-based modulation to regulate global staining style and intensity. The benefit of this design is clear when comparing \emph{PAINT w/o $f_{global}$} with the full model. Adding $f_{global}$ improves SSIM from 0.7128 to 0.7321 (+0.0193) and PSNR from 30.7752 to 31.2594 (+0.4842), and reduces LPIPS from 0.2493 to 0.2402 (-0.0091). This indicates that global conditioning helps maintain consistent staining intensity and overall appearance.

{\noindent\bf{Effect of Multi-scale Strategy.}}
The multi-scale strategy in PAINT is implemented via residual quantization to progressively decompose histological structures from coarse to fine scales. Its impact is evident when comparing \emph{PAINT w/o Multi-scale} to the full model, where enabling multi-scale modeling improves PSNR by 4.46 (26.7970$\rightarrow$31.2594) and reduces LPIPS by 0.13 (0.3718$\rightarrow$0.2402). These improvements highlight the benefit of multi-scale decomposition in capturing complex histological patterns and supporting robust high-resolution synthesis with enhanced perceptual quality.

{\noindent\bf{Effect of Registration.}}
To evaluate the necessity of explicit H\&E--IHC alignment, we remove the registration module. The \emph{w/o Registration} variant performs worse (SSIM: 0.6726; PSNR: 29.6429; LPIPS: 0.3104), while enabling registration improves SSIM to 0.7321 (+0.0595) and PSNR to 31.2594 (+1.6165), and reduces LPIPS to 0.2402 (-0.0702). These results indicate that correcting spatial misalignment reduces supervision noise and is a prerequisite for stable learning of cross-modal correspondence.

\begin{table*}[t]
\centering
\caption{Quantitative comparison across four IHC datasets: MIST-ER, MIST-PR, MIST-HER2, and MIST-Ki67. \textbf{Bold} indicates best results, and \underline{underline} indicates second-best results.}

\renewcommand{\arraystretch}{1.1}
\setlength{\tabcolsep}{2.2pt}

\begin{tabular}{c ccc ccc ccc ccc}
\toprule
& \multicolumn{3}{c}{\textbf{MIST-ER}}
& \multicolumn{3}{c}{\textbf{MIST-PR}}
& \multicolumn{3}{c}{\textbf{MIST-HER2}}
& \multicolumn{3}{c}{\textbf{MIST-Ki67}} \\
\cmidrule(lr){2-4}
\cmidrule(lr){5-7}
\cmidrule(lr){8-10}
\cmidrule(lr){11-13}

\textbf{Method}
& \textbf{SSIM} $\uparrow$ & \textbf{PSNR} $\uparrow$ & \textbf{LPIPS} $\downarrow$
& \textbf{SSIM} $\uparrow$ & \textbf{PSNR} $\uparrow$ & \textbf{LPIPS} $\downarrow$
& \textbf{SSIM} $\uparrow$ & \textbf{PSNR} $\uparrow$ & \textbf{LPIPS} $\downarrow$
& \textbf{SSIM} $\uparrow$ & \textbf{PSNR} $\uparrow$ & \textbf{LPIPS} $\downarrow$ \\
\midrule
Pix2pix~\cite{pix2pix}
& 0.1802 & 15.2513 & 0.5249
& 0.1817 & 15.3299 & 0.5168
& 0.1580 & 15.1582 & 0.5253
& 0.1734 & 15.0570 & 0.5192 \\
CycleGAN~\cite{cyclegan}
& 0.2183 & 14.7145 & 0.5065
& 0.2201 & 14.8420 & 0.5135
& 0.1897 & 14.6820 & 0.5185
& 0.2067 & 14.9717 & 0.5107 \\
CUT~\cite{cut}
& 0.2055 & 14.4950 & 0.5100
& 0.2101 & 14.5126 & 0.5075
& 0.1668 & 13.7156 & 0.5337
& 0.2071 & 14.8732 & 0.5034 \\
ASP~\cite{ASP-MIST}
& \textbf{0.2209} & 15.4460 & \underline{0.4994}
& 0.2113 & \underline{15.3521} & \underline{0.4889}
& 0.1915 & \underline{15.5460} & \underline{0.5021}
& 0.2061 & \underline{15.7310} & \underline{0.4995} \\
PPT~\cite{ppt}
& 0.2061 & \underline{15.5203} & 0.5373
& 0.2114 & 15.1567 & 0.5245
& 0.1791 & 14.5031 & 0.5437
& 0.1987 & 14.5261 & 0.5268 \\
PSPStain~\cite{pspstain}
& 0.2141 & 14.4162 & 0.5116
& 0.2100 & 14.7608 & 0.5027
& 0.1797 & 14.2899 & 0.5280
& 0.2129 & 15.2507 & 0.5010 \\
HistDiST~\cite{histdist}
& 0.1943 & 14.1084 & 0.5918
& 0.2000 & 13.8957 & 0.5957
& 0.1792 & 13.8214 & 0.6013
& 0.1999 & 14.3815 & 0.5912 \\
SynDiff~\cite{syndiff}
& 0.2100 & 14.8838 & 0.5115
& \underline{0.2267} & 14.4295 & 0.5109
& \underline{0.1920} & 13.9540 & 0.5320
& \textbf{0.2176} & 15.1106 & 0.5102 \\
PAINT-s
& \underline{0.2205} & \textbf{15.8477} & \textbf{0.4724}
& \textbf{0.2326} & \textbf{16.3104} & \textbf{0.4741}
& \textbf{0.1983} & \textbf{15.8843} & \textbf{0.4865}
& \underline{0.2161} & \textbf{16.0698} & \textbf{0.4700} \\
\bottomrule
\label{tab:mist_results}
\end{tabular}
\end{table*}

{\noindent\bf{Impact of Autoregressive Scale Configurations.}}
We finally investigate different scale schedules (bottom of Table~\ref{tab:ablation_PAINT}). A sparse progression (1, 2, 4, 8, 16) is suboptimal (SSIM: 0.6832; PSNR: 28.2692; LPIPS: 0.3322), implying insufficient intermediate refinement. Using dense scales from extremely coarse resolutions (1--16 dense) performs better (SSIM: 0.7236; PSNR: 31.0121; LPIPS: 0.2466) but still lags behind the proposed setting. Specifically, PAINT (8--16 dense) achieves SSIM = 0.7321, PSNR = 31.2594, and LPIPS = 0.2402, showing gains of +0.0085 SSIM, +0.2473 PSNR, and -0.0064 LPIPS compared to the second best schedule (1--16 dense). This suggests that initializing with the 3S-Map at mid-resolution and progressively refining with dense scales strikes the best balance between stability and detail recovery.

\section{Conclusion}
\label{sec:Conclusion}
In this paper, we presented PAINT, a pathology-aware visual autoregressive framework that rethinks virtual IHC as a structure-conditioned generation task rather than a deterministic translation problem. By addressing the inherently underdetermined relationship between H\&E morphology and IHC molecular expression, PAINT effectively mitigates the structural hallucinations prevalent in current approaches. A key innovation of our method is the 3S-Map, which enforces dual constraints in both latent and image spaces. This design effectively creates a morphology-grounded structural prior, preventing spatial drift and ensuring that generated molecular signals remain anatomically coherent with the underlying tissue architecture. Comprehensive evaluations on the IHC4BC and MIST datasets demonstrate PAINT's superiority over SOTA methods in generating high-fidelity, structurally accurate, and clinically viable staining patterns. While the multi-scale quantization inherent to VQ-VAE frameworks introduces a slight information bottleneck regarding fine-grained reconstruction, our results confirm that structure-guided autoregressive modeling offers a robust and principled direction for the future of computational pathology.

\bibliographystyle{IEEEtran}  
\bibliography{references}

\end{document}